\documentclass[conference]{IEEEtran}
\usepackage{amsmath,amsfonts}
\usepackage{graphicx}
\usepackage{booktabs}
\usepackage{multirow}
\usepackage[hidelinks,breaklinks=true]{hyperref}
\usepackage{placeins}

\title{Quantization Robustness to Input Degradations\\
for Object Detection}

\author{
\IEEEauthorblockN{Toghrul Karimov\IEEEauthorrefmark{1},
Hassan Imani\IEEEauthorrefmark{2},
Allan Kazakov\IEEEauthorrefmark{3}}

\IEEEauthorblockA{\IEEEauthorrefmark{1}Bahcesehir University, Baku, Azerbaijan}
\IEEEauthorblockA{\IEEEauthorrefmark{2}\IEEEauthorrefmark{3}Bahcesehir University, Istanbul, Turkey}

\IEEEauthorblockA{
\texttt{\{toghrul.karimov, allan.kazakov\}@bahcesehir.edu.tr} \\
\texttt{hassan.imani@bau.edu.tr}
}
}

\begin{document}
\maketitle

\begin{abstract}
Post-training quantization (PTQ) is crucial for deploying efficient object detection models, like YOLO, on resource-constrained devices. However, the impact of reduced precision on model robustness to real-world input degradations such as noise, blur, and compression artifacts is a significant concern. This paper presents a comprehensive empirical study evaluating the robustness of YOLO models (nano to extra-large scales) across multiple precision formats: FP32, FP16 (TensorRT), Dynamic UINT8 (ONNX), and Static INT8 (TensorRT). We introduce and evaluate a degradation-aware calibration strategy for Static INT8 PTQ, where the TensorRT calibration process is exposed to a mix of clean and synthetically degraded images. Models were benchmarked on the COCO dataset under seven distinct degradation conditions (including various types and levels of noise, blur, low contrast, and JPEG compression) and a mixed-degradation scenario. Results indicate that while Static INT8 TensorRT engines offer substantial speedups (\(\sim\)1.5--3.3x) with a moderate accuracy drop (\(\sim\)3--7\% mAP50-95) on clean data, the proposed degradation-aware calibration did not yield consistent, broad improvements in robustness over standard clean-data calibration across most models and degradations. A notable exception was observed for larger model scales under specific noise conditions, suggesting model capacity may influence the efficacy of this calibration approach. These findings highlight the challenges in enhancing PTQ robustness and provide insights for deploying quantized detectors in uncontrolled environments. All code and evaluation tables are available at \url{https://github.com/AllanK24/QRID}.
\end{abstract}


\section{Introduction}

Deep Neural Networks (DNNs), especially models like YOLO \cite{yolo}, have revolutionized object detection. However, their computational demands often hinder deployment on resource-constrained edge devices \cite{edge}. Model quantization, particularly Post-Training Quantization (PTQ), offers a compelling solution by converting FP32 models to lower-precision formats like INT8, significantly reducing size and improving inference speed without costly retraining \cite{quant}. 

While PTQ enhances efficiency, its effect on model robustness against real-world input degradations—such as noise, blur, low contrast, and compression artifacts—is a critical but less explored area \cite{robustness}. A primary concern is whether quantization, optimized for clean data, makes models more brittle in uncontrolled environments \cite{quant}.

This paper investigates this challenge by comprehensively evaluating the robustness of YOLO12 models (nano to extra-large scales) across multiple precision formats: FP32, FP16, Dynamic UINT8 (via ONNX Runtime), and Static INT8 (via NVIDIA TensorRT). We specifically examine how these different precision models perform on the COCO dataset \cite{coco} when subjected to a range of synthetically generated input degradations.

Furthermore, we propose and evaluate a degradation-aware calibration strategy for Static INT8 PTQ within the TensorRT framework. This approach modifies the standard calibration process by utilizing a dataset comprising a 50/50 mix of clean and degraded images, hypothesizing that this will lead to INT8 models with enhanced resilience to such input imperfections.

Our main contributions are as follows:
\begin{itemize}
\item A multi-scale empirical analysis of quantization's impact on YOLO12 robustness to common image degradations.
\item The evaluation of a novel degradation-aware static calibration method.
\item Quantitative insights into the accuracy, speed, and robustness trade-offs, providing practical guidance for deploying efficient yet reliable object detectors in real-world scenarios.
\end{itemize}

The study reveals that while Static INT8 offers significant speedups, our proposed mixed-data calibration did not consistently improve robustness over standard clean-data calibration, except for potential benefits against noise in larger models.

\section{Related Work}

Deploying deep learning models for real-time object detection on resource-constrained edge devices presents significant challenges due to the computational and memory demands of state-of-the-art architectures \cite{han2016compression}. This has driven extensive research into model compression and acceleration techniques.

\subsection{Efficient Object Detection with YOLO}

The "You Only Look Once" (YOLO) family of object detectors, introduced by Redmon et al. \cite{yolo}, revolutionized real-time object detection by framing it as a single regression problem. Subsequent iterations, such as YOLOv4 \cite{yolov4} and the widely adopted versions from Ultralytics like YOLOv5 and YOLOv8 \cite{ultralytics}, have consistently pushed the Pareto frontier of speed and accuracy. These models are frequently targeted for edge deployment due to their efficient architectures, making them pertinent subjects for quantization studies.

\subsection{Post-Training Quantization for Neural Networks}

Model quantization is a key technique for reducing model size, inference latency, and power consumption by representing weights and/or activations with lower-precision data types, typically converting from 32-bit floating-point (FP32) to 8-bit integers (INT8) \cite{quant1,krishnamoorthi2018quantizing}. Post-Training Quantization (PTQ) is particularly appealing as it does not require retraining or access to the original training pipeline.

Two main PTQ approaches exist: Dynamic Quantization, where activation ranges are determined at runtime, incurring some computational overhead but simplifying application \cite{quant2}; and Static Quantization, where activation ranges are pre-determined by calibrating the model on a small, representative dataset. Static PTQ, often implemented using the Quantize-Dequantize (QDQ) format for ONNX models or through tools like NVIDIA TensorRT's calibration process for engine generation, typically offers superior latency improvements \cite{tensorrt}. While Quantization-Aware Training (QAT) can yield higher accuracy by simulating quantization during training \cite{quant1}, this study focuses on PTQ due to its practical advantages for pre-trained models.

\subsection{Quantization and Robustness to Input Degradations}

A critical concern for real-world deployment is how quantization affects a model's robustness to common input degradations like noise, blur, or compression artifacts. Several studies have indicated that the reduction in numerical precision can make quantized models more sensitive to such perturbations compared to their FP32 counterparts \cite{robustness,dodge2017distortions}. For instance, Shafiee et al. \cite{shafiee2021lowprecision} explored the brittleness of quantized networks, and some works suggest that quantization can amplify the effects of input noise.

Attempts to mitigate this robustness gap have included specialized training techniques or architectural modifications \cite{qat}. However, enhancing the robustness of post-training quantized models, particularly for widely used architectures like YOLO deployed via optimized inference engines like TensorRT, remains an active area. While general PTQ calibration aims to preserve accuracy on clean data, standard calibration procedures typically use only high-quality images, potentially leading to suboptimal performance when the model encounters degraded inputs. Few studies have systematically evaluated the impact of diverse degradations across multiple scales of YOLO models under common PTQ scenarios (TensorRT INT8, ONNX Dynamic UINT8) or explored simple yet practical modifications to the PTQ calibration data itself to directly address robustness. This paper aims to fill this gap by empirically analyzing the robustness of quantized YOLO12 models and evaluating a degradation-aware calibration strategy designed to improve resilience to common input corruptions.

\section{Methodology}

This study systematically evaluated the impact of post-training quantization on the robustness of five YOLO12 model scales (nano, small, medium, large, and extra-large) to common image degradations. All experiments were conducted on an NVIDIA RTX 2070 GPU using the COCO 2017 dataset for both calibration and validation \cite{coco}.

The baseline FP32 models were converted into several precision formats: FP16 (TensorRT engine) \cite{tensorrt}, Dynamic UINT8 (ONNX model via ONNX Runtime) \cite{onnx}, and Static INT8 (TensorRT engine) \cite{tensorrt}. 

For Static INT8, two calibration strategies were investigated:
\begin{enumerate}
\item Standard calibration using a dataset of 1000 clean images sampled from COCO train2017 \cite{coco_train}
\item A proposed degradation-aware calibration using a dataset of 1000 images comprising a 50/50 mix of clean and synthetically degraded samples (derived from the same clean sample pool) \cite{qrid}
\end{enumerate}

Dynamic quantization (targeting weights of operations like Conv/MatMul) was applied to ONNX models using ONNX Runtime, without requiring a separate calibration dataset for activations.

To assess robustness, the full COCO val2017 set was used to generate multiple distinct validation datasets. Each set featured a specific synthetic degradation applied uniformly to all images using the Albumentations library \cite{albumentations}.

The parameters for these degradations were chosen to represent plausible real-world imperfections:
\begin{itemize}
\item Gaussian Noise: standard deviation range of (10/255, 30/255) for Low and (35/255, 55/255) for Medium
\item Gaussian Blur: kernel size limits of (3, 5) for Low and (7, 11) for Medium
\item Low Contrast: contrast limit of (-0.6, -0.3)
\item Heavy JPEG Compression was applied with a quality range of (20, 45).
\end{itemize}

This pool of degradations was also used for the Mixed Degradation validation set, in which 50\% of the validation images were randomly subjected to one of these transformations.

All models were benchmarked with a batch size of 1. Performance was primarily evaluated using standard COCO metrics (mAP50-95, mAP50) and inference latency (ms/image), obtained via the Ultralytics \texttt{model.val()} framework \cite{ultralytics}.

Robustness was quantified by analyzing the absolute and relative mAP drops when transitioning from clean to degraded validation data, with a particular focus on comparing the efficacy of the clean versus degradation-aware static calibration strategies \cite{qrid}.

\section{Experiments and Results}

This section presents the empirical evaluation of YOLO12 model variants (nano to extra-large) across different precision formats (FP32, FP16, Dynamic UINT8, and Static INT8 with clean/mixed calibration) on the COCO val2017 dataset \cite{coco}, both in its original clean form and under various synthetic degradations.

\subsection{Baseline Performance on Clean COCO Validation Data}

We first established baseline performance on the clean COCO val2017 dataset. Table~\ref{tab:baseline_accuracy} summarizes key accuracy metrics (mAP50-95 and mAP50), while Table~\ref{tab:latency} details the corresponding inference latencies and FPS.

\begin{table*}[!t]
\small
\centering
\caption{Baseline Performance on COCO val2017 set}
\label{tab:baseline_accuracy}
\begin{tabular}{cccccc}
\toprule
Model Scale & Precision & Calibration & mAP50-95 & mAP50 \\
\midrule
N & FP32 (TRT) & N/A & 0.4047 & 0.5686 \\
N & FP16 (TRT) & N/A & 0.4044 & 0.5685 \\
N & Dynamic UINT8 (ONNX) & N/A & 0.4047 & 0.5686 \\
N & Static INT8 (TRT) & Clean & 0.3325 & 0.4787 \\
N & Static INT8 (TRT) & Mixed & 0.3327 & 0.4789 \\
\midrule
S & FP32 (TRT) & N/A & 0.4763 & 0.6462 \\
S & FP16 (TRT) & N/A & 0.4763 & 0.6464 \\
S & Dynamic UINT8 (ONNX) & N/A & 0.4763 & 0.6462 \\
S & Static INT8 (TRT) & Clean & 0.4114 & 0.5670 \\
S & Static INT8 (TRT) & Mixed & 0.4114 & 0.5670 \\
\midrule
M & FP32 (TRT) & N/A & 0.5226 & 0.6943 \\
M & FP16 (TRT) & N/A & 0.5222 & 0.6939 \\
M & Dynamic UINT8 (ONNX) & N/A & 0.5226 & 0.6943 \\
M & Static INT8 (TRT) & Clean & 0.4853 & 0.6517 \\
M & Static INT8 (TRT) & Mixed & 0.4853 & 0.6517 \\
\midrule
L & FP32 (TRT) & N/A & 0.5347 & 0.7049 \\
L & FP16 (TRT) & N/A & 0.5345 & 0.7050 \\
L & Dynamic UINT8 (ONNX) & N/A & 0.5347 & 0.7049 \\
L & Static INT8 (TRT) & Clean & 0.5079 & 0.6802 \\
L & Static INT8 (TRT) & Mixed & 0.5079 & 0.6803 \\
\midrule
X & FP32 (TRT) & N/A & 0.5514 & 0.7194 \\
X & FP16 (TRT) & N/A & 0.5513 & 0.7193 \\
X & Dynamic UINT8 (ONNX) & N/A & 0.5514 & 0.7194 \\
X & Static INT8 (TRT) & Clean & 0.5202 & 0.6881 \\
X & Static INT8 (TRT) & Mixed & 0.5032 & 0.6659 \\
\bottomrule
\end{tabular}
\end{table*}

\begin{table*}[!t]
\small
\centering
\caption{Latency (ms) on CLEAN data}
\label{tab:latency}
\begin{tabular}{cccccc}
\toprule
Model Scale & Precision & Calibration & Latency (ms) & FPS \\
\midrule
N & FP32 (TRT) & N/A & 3.8 & 262.6 \\
N & FP16 (TRT) & N/A & 2.2 & 448.8 \\
N & Dynamic UINT8 (ONNX) & N/A & 6.5 & 153.2 \\
N & Static INT8 (TRT) & Clean & 2.6 & 380.9 \\
N & Static INT8 (TRT) & Mixed & 2.7 & 370.8 \\
\midrule
S & FP32 (TRT) & N/A & 8.6 & 116.2 \\
S & FP16 (TRT) & N/A & 3.5 & 288.3 \\
S & Dynamic UINT8 (ONNX) & N/A & 12.1 & 82.9 \\
S & Static INT8 (TRT) & Clean & 4.0 & 252.9 \\
S & Static INT8 (TRT) & Mixed & 4.0 & 248.6 \\
\midrule
M & FP32 (TRT) & N/A & 20.5 & 48.7 \\
M & FP16 (TRT) & N/A & 6.3 & 157.5 \\
M & Dynamic UINT8 (ONNX) & N/A & 25.5 & 39.2 \\
M & Static INT8 (TRT) & Clean & 6.9 & 145.3 \\
M & Static INT8 (TRT) & Mixed & 6.7 & 149.5 \\
\midrule
L & FP32 (TRT) & N/A & 30.3 & 33.0 \\
L & FP16 (TRT) & N/A & 10.2 & 98.2 \\
L & Dynamic UINT8 (ONNX) & N/A & 36.9 & 27.1 \\
L & Static INT8 (TRT) & Clean & 11.5 & 87.0 \\
L & Static INT8 (TRT) & Mixed & 11.8 & 84.5 \\
\midrule
X & FP32 (TRT) & N/A & 61.3 & 16.3 \\
X & FP16 (TRT) & N/A & 18.2 & 55.0 \\
X & Dynamic UINT8 (ONNX) & N/A & 67.5 & 14.8 \\
X & Static INT8 (TRT) & Clean & 18.4 & 54.5 \\
X & Static INT8 (TRT) & Mixed & 18.5 & 54.0 \\
\bottomrule
\end{tabular}
\end{table*}

Observations from Table~\ref{tab:baseline_accuracy} and Table~\ref{tab:latency} reveal clear trade-offs. As expected, increasing model scale from Nano (n) to Extra-Large (x) consistently resulted in higher baseline accuracy (e.g., YOLO12x FP32 mAP50-95 of 0.5514 vs. YOLO12n FP32 of 0.4047, Table~\ref{tab:baseline_accuracy}) but also led to substantially increased inference latency (e.g., 61.3 ms vs. 3.8 ms respectively, Table~\ref{tab:latency}).

Comparing precision formats for a given model scale:
\begin{itemize}
\item FP16 (TensorRT) demonstrated negligible accuracy loss compared to FP32 \cite{tensorrt} (Table~\ref{tab:baseline_accuracy}) while providing significant speedups (e.g., YOLO12n latency reduced from 3.8 ms to 2.2 ms, Table~\ref{tab:latency}).
\item Dynamic UINT8 (ONNX), as shown in Table~\ref{tab:baseline_accuracy}, matched the FP32 models in mAP. However, Table~\ref{tab:latency} indicates it consistently exhibited higher latency (lower FPS) than the FP32 TensorRT baseline (e.g., 6.5 ms for YOLO12n Dynamic UINT8 vs. 3.8 ms for FP32 TRT). This is attributed to runtime quantization overheads within the ONNX Runtime framework, and its inclusion primarily serves as an accuracy reference for uncalibrated PTQ.
\item Static INT8 (TensorRT) provided the greatest latency reductions \cite{tensorrt} (Table~\ref{tab:latency}), achieving, for instance, 2.6 ms for YOLO12n (Clean Calib) compared to 3.8 ms for FP32. This efficiency, however, came with a notable drop in accuracy on clean data (Table~\ref{tab:baseline_accuracy}), with mAP50-95 reductions ranging from approximately 0.031 for YOLO12x (Clean Calib) to 0.072 for YOLO12n (Clean Calib) compared to their FP32 counterparts.
\item The choice of calibration data (Clean vs. Mixed) for Static INT8 models had a minimal effect on both accuracy (Table~\ref{tab:baseline_accuracy}) and speed (Table~\ref{tab:latency}) when evaluated on this clean dataset, with only the YOLO12x model showing a slightly more pronounced mAP drop with mixed calibration.
\end{itemize}
These baseline characteristics are crucial for contextualizing the subsequent robustness analysis under degraded input conditions.

\subsection{Robustness to Input Degradations and Evaluation of Calibration Strategies}

To assess model robustness and the efficacy of different quantization strategies under imperfect conditions, all model configurations were evaluated on a suite of validation datasets. These sets were generated from the COCO val2017 images \cite{coco}, each featuring a specific synthetic degradation: Gaussian Noise (Low, Medium), Gaussian Blur (Low, Medium), Low Contrast, or Heavy JPEG Compression. Additionally, a general Mixed Degradation scenario was tested. The impact of these degradations is quantified by the relative drop in mAP50-95 performance compared to each model's own clean data baseline (Table~\ref{tab:baseline_accuracy}) \cite{robustness}.

Table~\ref{tab:robustness} details the relative mAP50-95 drop percentages for a selection of these degradations: Medium Blur, Low Noise, Medium Noise, and the Mixed Degradation set. Degradations such as Low Blur, Low Contrast, and Heavy JPEG Compression, which generally induced minimal relative mAP drops across most configurations (typically <2\%), are primarily summarized textually for brevity.

\begin{table*}[!t]
\small
\centering
\caption{Relative Drop mAP50-95 (\%)}
\label{tab:robustness}
\begin{tabular}{ccccccc}
\toprule
Model Scale & Precision & Calibration & Blurry Medium & Noisy Low & Noisy Medium & Mixed Degrad \\
\midrule
N & FP32 & N/A & 13.10\% & 24.50\% & 59.30\% & 7.80\% \\
N & FP16 & N/A & 13.10\% & 24.50\% & 59.30\% & 7.80\% \\
N & Dynamic UINT8 & N/A & 13.10\% & 24.50\% & 59.30\% & 7.80\% \\
N & Static INT8 & Clean & 13.20\% & 29.40\% & 60.60\% & 7.60\% \\
N & Static INT8 & Mixed & 13.20\% & 29.30\% & 60.70\% & 7.60\% \\
\midrule
S & FP32 & N/A & 12.40\% & 18.50\% & 47.10\% & 6.20\% \\
S & FP16 & N/A & 12.50\% & 18.60\% & 47.10\% & 6.30\% \\
S & Dynamic UINT8 & N/A & 12.40\% & 18.50\% & 47.10\% & 6.20\% \\
S & Static INT8 & Clean & 11.50\% & 23.60\% & 47.70\% & 6.50\% \\
S & Static INT8 & Mixed & 11.50\% & 23.60\% & 47.60\% & 6.50\% \\
\midrule
M & FP32 & N/A & 12.30\% & 13.30\% & 31.60\% & 5.30\% \\
M & FP16 & N/A & 12.30\% & 13.20\% & 31.60\% & 5.30\% \\
M & Dynamic UINT8 & N/A & 12.30\% & 13.30\% & 31.60\% & 5.30\% \\
M & Static INT8 & Clean & 11.00\% & 17.20\% & 36.10\% & 5.50\% \\
M & Static INT8 & Mixed & 11.00\% & 22.00\% & 36.10\% & 5.50\% \\
\midrule
L & FP32 & N/A & 11.60\% & 12.00\% & 29.70\% & 4.60\% \\
L & FP16 & N/A & 11.70\% & 12.00\% & 29.60\% & 4.60\% \\
L & Dynamic UINT8 & N/A & 11.60\% & 12.00\% & 29.70\% & 4.60\% \\
L & Static INT8 & Clean & 11.70\% & 18.40\% & 30.00\% & 4.70\% \\
L & Static INT8 & Mixed & 11.60\% & 12.90\% & 30.00\% & 4.60\% \\
\midrule
X & FP32 & N/A & 11.90\% & 11.10\% & 27.20\% & 4.20\% \\
X & FP16 & N/A & 11.90\% & 11.20\% & 27.30\% & 4.30\% \\
X & Dynamic UINT8 & N/A & 11.90\% & 11.10\% & 27.20\% & 4.20\% \\
X & Static INT8 & Clean & 15.10\% & 12.50\% & 34.70\% & 7.80\% \\
X & Static INT8 & Mixed & 12.20\% & 9.60\% & 28.10\% & 4.70\% \\
\bottomrule
\end{tabular}
\end{table*}

Key observations from this robustness analysis (Table~\ref{tab:robustness} and overall data) are as follows: Gaussian noise (Low and Medium) consistently induced the most severe performance degradation across all models and precisions. For instance, under Medium Noise, relative mAP drops for YOLO12n Static INT8 exceeded 60\%, while for YOLO12x FP32, the drop was around 27\%. Medium Blur also caused significant degradation, with relative drops typically ranging from 11-15\% depending on model scale and precision. In contrast, Low Contrast and Heavy JPEG Compression had a considerably milder impact, with most models experiencing less than a 2\% relative drop, indicating high resilience to these types. The Mixed Degradation set, reflecting an average of effects, resulted in moderate overall performance decreases (typically 4-8\% relative drops).

Comparing precision formats based on Table~\ref{tab:robustness}, the relative robustness patterns of FP32, FP16, and Dynamic UINT8 models were generally similar for a given model scale under the shown degradations. Static INT8 models (both clean and mixed calibration) often exhibited different characteristics; for example, under Medium Blur, the Static INT8 relative drops were sometimes comparable or slightly smaller than FP32. However, under both Low and Medium Noise, Static INT8 models consistently demonstrated increased sensitivity (larger relative drops) compared to the FP32/FP16 models.

The central investigation into degradation-aware (mixed) calibration for Static INT8 models, as seen in Table~\ref{tab:robustness}, showed that for most model scales and the presented degradations, this strategy yielded robustness performance remarkably similar to standard clean data calibration. Notable exceptions where mixed calibration appeared beneficial were observed primarily for the YOLO12x model under Noisy Medium conditions (28.1\% drop for mixed vs. 34.7\% for clean) and also under Noisy Low (9.6\% vs 12.5\%) and the Mixed Degradation set (4.7\% vs 7.8\%). However, these gains were not consistently replicated across smaller model scales or other degradation types. Inference latencies (Table~\ref{tab:latency}) remained largely unaffected by the input image degradations themselves.

\FloatBarrier

\section{Conclusion and Discussion}

This paper investigated the impact of post-training quantization (PTQ) on the robustness of YOLO12 object detection models (nano to extra-large scales) when subjected to common image degradations. Driven by the demand for efficient edge deployment, we compared standard FP32 and FP16 TensorRT engines against Dynamic UINT8 (ONNX Runtime), standard Static INT8 (TensorRT, clean data calibration), and a proposed degradation-aware Static INT8 variant calibrated with a 50/50 mix of clean and synthetically degraded images. Performance was assessed using mAP metrics and inference latency on the COCO val2017 dataset under seven distinct degradation conditions and a mixed-degradation scenario.

Our findings highlight key trade-offs. As expected, larger model scales yielded higher accuracy but with increased latency. FP16 precision offered a compelling balance, significantly improving speed over FP32 with negligible accuracy loss. Static INT8 quantization provided the greatest inference speedup (\(\sim\)1.5--3.3x vs. FP32) but incurred a baseline accuracy drop of \(\sim\)3--7\% absolute mAP50-95, with smaller models being more affected. Dynamic UINT8, while maintaining FP32-level accuracy, was slower than the optimized FP32 TensorRT baseline due to its runtime overhead, serving primarily as an uncalibrated PTQ accuracy reference.

Regarding robustness, all models experienced performance degradation under corruptions, with Gaussian noise proving exceptionally detrimental (e.g., \(>59\%\) relative mAP drop for YOLO12n Static INT8), indicating a significant vulnerability across all tested precisions. Other degradations like blur induced moderate drops, while low contrast and JPEG compression had minimal impact. Interestingly, quantized INT8 models were not uniformly more brittle than their FP32/FP16 counterparts; for degradations like blur and contrast, their relative performance drops were often comparable or even slightly smaller. However, Static INT8 models consistently showed increased sensitivity to the applied noise levels.

The central investigation into degradation-aware calibration revealed limited overall efficacy. While a notable robustness improvement against noise was observed for the largest model scale (YOLO12x), this benefit did not consistently extend to smaller models or other degradation types. For most scenarios, mixed calibration performed similarly to standard clean data calibration, suggesting that the specific implementation (50/50 data mix with TensorRT's default calibration) did not broadly enhance resilience. This underscores the challenge in designing universally effective PTQ robustness strategies.

Limitations of this study include its focus on the YOLO12 family and the COCO dataset, the specific set and severity of synthetic degradations, and reliance on the default TensorRT calibration mechanisms. Future work should explore more advanced or adaptive calibration algorithms, conduct layer-wise sensitivity analyses particularly for noise, investigate different clean-to-degraded data ratios for calibration, evaluate targeted mixed-precision strategies, and compare findings against Quantization-Aware Training (QAT) approaches to further advance the deployment of robust and efficient object detectors. In conclusion, while PTQ, especially Static INT8 via TensorRT, offers crucial efficiency gains, enhancing its robustness to diverse real-world degradations, particularly severe noise, remains an important challenge requiring more sophisticated solutions than simple mixed-data calibration for most model scales.

\section*{Acknowledgment}

This work is supported in part by the Scientific and Technological Research Council of Turkey (TUBITAK) under Project No. 124E099.

\clearpage
\bibliographystyle{IEEEtran}
\bibliography{references}

\end{document}